\documentclass{article}
\usepackage{cite}
\usepackage{amsmath,amssymb,amsfonts}
\usepackage{algorithmic}
\usepackage{graphicx}
\usepackage{textcomp}
\usepackage{xcolor}
\usepackage{tabularx}
\usepackage{longtable}
\def\BibTeX{{\rm B\kern-.05em{\sc i\kern-.025em b}\kern-.08em
    T\kern-.1667em\lower.7ex\hbox{E}\kern-.125emX}}
\begin{document}


\textcolor{red}{Please note: This is a preprint version of a submitted conference paper before reviews:
“This work has been submitted to the IEEE for possible publication. Copyright may be transferred without notice, after which this version may no longer be accessible.”}

Authors: Martin Cooney, Eric Järpe, Alexey Vinel

Title: ``Robot Steganography''?: Opportunities and Challenges

Address: Center for Applied Intelligent Systems Research (CAISR), School of Information Technology, Halmstad University, P. O. Box 823, Kristian IV:s väg 3, 30118 Halmstad, Sweden

Contact: martin.daniel.cooney@gmail.com

Acknowledgments: Funding was received from the Swedish Knowledge Foundation through the SafeSmart Synergy project, ”Safety of Connected Intelligent Vehicles in Smart Cities”, which we gratefully acknowledge.

\title{``Robot Steganography''?: \\ Opportunities and Challenges\\} 
\date{\vspace{-5ex}}

 \maketitle

\begin{abstract}
Robots are being designed to communicate with people in various public and domestic venues in a perceptive, helpful, and discreet way.
Here, we use a speculative design approach to shine light on a new concept of \emph{robot steganography} (RS): that a robot could seek to help vulnerable populations by discreetly warning of potential threats:
We first identify some potentially useful scenarios for RS related to safety and security--concerns that are estimated to cost the world trillions of dollars each year--with a focus on two kinds of robots, an autonomous vehicle (AV) and a socially assistive humanoid robot (SAR).
Next, we propose that existing, powerful, computer-based steganography (CS) approaches can be adopted with little effort in new contexts (SARs), while also pointing out potential benefits of human-like steganography (HS): 
Although less efficient and robust than CS, HS represents a currently-unused form of RS that could also be used to avoid requiring a computer to receive messages, detection by more technically advanced adversaries, or a lack of alternative connectivity (e.g., if a wireless channel is being jammed).
This analysis also introduces some unique challenges of RS that arise from message generation, indirect perception, and effects of perspective.
For this, we explore some related theoretical and practical concerns for selecting carrier signals and generating messages, also making available some code and a video demo.
Finally, we confirm the basic feasibility of RS, that messages can be hidden in a robot’s behaviors, via a simplified user study.
The immediate implication is that RS could help to improve people's lives and mitigate some costly problems--suggesting the usefulness of further discussion, ideation, and consideration by designers.
\end{abstract}

\section{Introduction}

At the crossroads between human-robot interaction and secure communications, this design paper focuses on the emerging topic of "robot steganography" (RS), the hiding of messages by a robot.

Steganography, or message hiding, is a crucial security technique that can complement other approaches.
For example, encryption alone cannot prevent an adversary from detecting that a message is being sent.

Steganographic messages could also be sent by interactive robots, which are expected to play an increasingly useful role in the smart cities of the near future by conducting dull, dangerous, and dirty tasks, in a scalable, engaging, reliable, and perceptive way.
Here “robot” is defined generally as an embedded computing system, comprising sensors and actuators that afford some semi-autonomous, intelligent, or human-like qualities.
Thus, this definition includes systems that we might not normally think of as robots, such as autonomous vehicles (AVs), smart homes, and wearables (i.e., robots that we ride, live in, and wear), as well as more common tropes like humanoid robots (robots that excel at human-like communication).

Such robots exhibit qualities conducive for steganography:
\begin{itemize}
\item{{\bf Generality}. Since robots typically contain computers, existing computer-based steganography (CS) approaches can be used.}
\item{{\bf Multimodality}. Robots can generate various signals, from motions to sounds, that provide opportunities to hide messages.}
\item{{\bf Opacity}. Robots tend to be highly complex, such that most people do not understand how they work.}
\item{{\bf Nascency}. Robots are not yet generally common in everyday human environments due to their current level of technological readiness, which could allow for occasional odd behavior to be overlooked (plausible deniability).}
\end{itemize}

However, currently it is unclear how a robot can seek to accomplish good via steganography.
Thus, the goal of the current paper is to explore this gap, using a speculative scenario-building approach--which seeks to provoke thought by constructing concrete "memories" of a potential future reality via rapid ideation and discussion sessions~\cite{rasmussen2005narrative}.

The remainder of this paper is organized as follows: 
Section~\ref{section:related-work} discusses some related work, identifying gaps;
Section~\ref{section:motivation} describes two scenarios
related to outdoor and indoor robots, indicating
some unique challenges related to behavior generation and perception.
Some theoretical and practical engineering concerns, for selecting carrier signals and generating messages, are explored in ~\ref{section:carriers} and ~\ref{section:signal-generation}.
This leads
to a proof-of-concept implementation, that is
checked via a user study, as reported in ~\ref{section:user-study}. 
Finally, the results are discussed in Section~\ref{section:discussion}, along with ideas for next steps. 
Thereby, the aim is to stimulate thought about the possibilities for robots to help people in the near future.

\section{Related Work}
\label{section:related-work}

There is a long history of work in intelligent robotics that argues that some robots should be perceptive and capable of interacting without explicit commands in a contextually appropriate manner~\cite{schmidt2000implicit}--steganography approaches have been developed that could be adapted for such robots.

\subsection{Robot discretion}

Various approaches have been proposed to facilitate rich, mixed-initiative interactions, from leveraging familiar interfaces to implicit interaction guidelines (e.g., interaction analogues that allow an interacting person to preserve their attention and that do not presume without precedent~\cite{ju2015design}). 
Along similar lines, researchers have recently proposed that robots that interact with humans should not always single-mindedly reveal truth, but will need to “lie” in various situations, to provide good service~\cite{wagner2016lies, isaac2017robots}.
For example, a robot asked by its owner about their weight might not wish to respond, "Yes, you are very fat".

Toward this goal, work has started to identify relevant behavioral approaches and concerns:
Wagner et al. reported on applying \emph{interdependence theory}, suggesting that stereotypes can be used to initially estimate the cost, value, and estimated success rate of lying~\cite{wagner2016lies}.
Isaac and colleagues indicated the usefulness of \emph{theory of mind} to allow robots to detect ulterior motives, to avoid manipulation by humans with bad intentions~\cite{isaac2017robots}.
Additionally, Gonzalez-Billandon et al. developed a system to detect human lies based on eye movements, response times, and eloquence, also verifying that robots were lied to in a similar way as humans~\cite{10.3389/frobt.2019.00064}.
Such work has formed a basis for robots to interact more effectively via discreet communication.

We believe that for similar reasons, not just false utterances, but also an ability to send secret messages to the right recipient via steganography could be useful.

\subsection{Traditional steganography}

Here, we propose that current steganography can be considered as comprising two broad categories: human and computer steganography.

\subsubsection{Human steganography.} Humans have used steganography at least since the age of Ancient Greece to warn of threats and ask for help, using a variety of audiovisual signals~\cite{petitcolas1999information}.
For example, steganography has been used to indicate mistreatment in war by POWs who blinked in Morse code or used rude gestures that their captors did not recognize.\footnote{https://www.archives.gov/exhibits/eyewitness/html.php?section=8}$^{,}$\footnote{https://www.damninteresting.com/the-seizing-of-the-pueblo/}
Various signals have also been proposed for reporting domestic violence, such as by asking for “Angela” or a “Minotaur” at a bar, drawing a black dot on one’s hand with mascara\footnote{https://www.bbc.com/news/blogs-trending-34326137}, using a red pen instead of a black pen at a clinic\footnote{https://www.lgbtqnation.com/2020/01/clinics-ingenious-way-help-domestic-violence-victims-sweeping-web/}, or using a hand gesture\footnote{https://www.bodyandsoul.com.au/health/health-news/a-secret-hand-signal-showing-someone-is-a-victim-of-domestic-violence-is-going-viral/news-story/3b6ab5c0cd03052fba02d880312c7d3b}.
In general, such signals are also common in everyday media tropes, from gestures such as putting "bunny ears" behind someone's head when a photo is taken, to facial expressions behind someone's back, or using a bird call to signal an ally without alerting adversaries.

One recent study has started to venture into this apparently mostly uncharted domain, examining how an underwater vehicle could mimic animal sounds~\cite{jia2018bio}.
Yet, such studies exploring how robots could conduct steganography in a human-like way (hereafter HS) appear to be strikingly
rare--possibly due to the nascent state of robotic technology, as well as some salient advantages of computer-based steganography (CS).

\subsubsection{Computer-based steganography (CS).} The development of computers led to new possibilities for highly efficient and robust steganography, which typically involves small changes being made to little-used, redundant parts of a digital carrier signal.
For example, least significant bits (LSB), parity bits, or certain frequencies can be used, in a carrier such as digital text, visual media (image, video), audio (music, speech, sounds), or network communications (communicated frames/data packets)~\cite{zielinska2014trends}.
In particular, the application of the latter to robots has started to be explored; for example, de Fuentes and colleagues investigated steganography in Vehicular Ad hoc Networks (VANETs)~\cite{de2014applying}.
However, there is a gap related to the "big picture" of how robots in general could engage in steganography, which requires a combination of technical and design perspectives.

In our previous work, we have conducted a user study to explore the robustness of a proposed steganography method~\cite{jarpe2021velody}, and reported in a short paper on some initial ideas regarding vehicular steganography~\cite{cooney2021}. 
The novel contribution of the current paper, which extends the latter, is in using a speculative prototyping approach to explore the "big picture" for RS, including useful scenarios and how the challenges that emerge could be overcome, as well as basic feasibility via a user study.

\section{Methods}
\label{section:methods}

To explore the lay of the land, we adopted a \emph{speculative} design approach combining \emph{scenario-building} and \emph{prototyping}, to identify potential conceptual interactions from a design and technical perspective, and conduct a simplified user study.
“Speculative design” is a simplified, fictional, problem-finding approach, intended to open a portal to see how things could be in an alternative future reality and thereby provoke thought and stimulate discussion~\cite{dunne2013speculative}.
A core tool in the speculative toolbox is the “scenario”, an open-ended story that gives us “memories of the future”--by communicating visions in a concrete, easily-understood, relatable, and interesting way~\cite{rasmussen2005narrative}.
However, imagined scenarios alone might miss capturing what could happen in the real world; for this, “prototyping” offers a way to avoid prohibitively time-consuming and expensive manufacturing of full systems, by balancing speed of investigation with accuracy of insights, in line with the maxim, “Fail often, fail fast, fail cheap"~\cite{engelberg2002framework}.

Thus, the notion of RS was explored first through rapid ideation sessions and scenario building.
Discussion within the group raised a number of questions:
\begin{itemize}
  \item {\bf Big Picture}. How might a robot help people by sending secret messages?
  \item {\bf Carriers}. What signals can be used to hide messages?
  \item {\bf Signal Generation}. What theoretical and practical concerns exist?
\end{itemize}
Furthermore, from the different kinds of robot we had identified based on our generic definition--robots that we ride, play with, live in, and wear--two main kinds of robot were identified as lenses to facilitate exploration of the questions: an \emph{autonomous vehicle (AV)} and \emph{a socially assistive humanoid robot (SAR)}.
The former is an outdoor robot with a focus on transport and movement, whereas the latter is an indoor robot with a focus on social communication (especially for healthcare). 
Both offer exciting possibilities for improving quality of life in interacting persons.
Although some might argue that an AV is not a robot, like various others (e.g., ~\cite{wang2020intelligent}), we see this as a kind of mobile robot that is very important to consider, also given the continual, high rate of development and vibrancy in this area, and the rate at which AVs are entering our cities and becoming an ever more practical reality.
(Various other questions and kinds of robots might exist, and the ones presented here merely act as a basis for initial exploration and are by no means the only possible options.)
 
\subsection{The "Big Picture"}
\label{section:motivation}

In line with the "How Might We" design method, the first question identified was phrased as: How might a robot help people by sending secret messages?

To address the question, brainstorming ideas were recorded without judgement, then blended and grouped into short written narrative scenarios.
In doing so, the aim was to capture a wide range of ideas in a small number of potentially high-value, plausible scenarios.
We stress that feasibility from the perspective of current technology was not used as a filter, given the speculative approach; i.e., our initial concern was not \emph{how} robot capabilities could be implemented (such as the rich recognition capabilities that willbe required) but \emph{what} could be useful.
This resulted in a total of eight scenarios (four per category).
The scenarios were then analyzed, yielding insight into some core themes: the kinds of problems that would be useful to design solutions for, commonalities, venues, interactive roles, cues to detect, and actions a robot could take, as well as some unique challenges.

Two example scenarios are presented below:

\emph{{\bf AV}. 
"Hey!" KITTEN, a large truck AV, inadvertently exclaimed. "Are you watching the road?"  
Its driver, Oscar, ignored KITTEN, speeding erratically down the crowded street near the old center of the city with its tourist area, market, station, and school, which were not on his regular route.
KITTEN was worried about Oscar, who had increasingly been showing signs of radicalization--meeting with extremists such as Mallory--and instability, not listening to various warnings related to medicine non-adherence, depression, and sleep deprivation.
But she wasn't completely sure if Oscar was currently dangerous or impaired, as his driving was always on the aggressive side; and, 
KITTEN didn't want to go to the police--if she were wrong, Oscar might lose his job. Or, even if she were right and the police didn't believe her, Oscar could get angry and try to bypass her security feature, or find a different car altogether, and then there would be no way to help anymore.
At the next intersection, KITTEN decided to use network steganography to send a quick "orange" warning to nearby protective infrastructure, comprising a monitoring system and anti-tire spikes that can be raised to prevent vehicles from crashing into crowds of pedestrians--while planning to execute an emergency brake and call for help if absolutely required.}

\emph{{\bf SAR}. 
"Howdy!" called Alice, the cleaning robot at the care center, as she entered Charlie's room. Her voice trailed off as she took in the odd scene in front of her: Charlie appeared agitated, and she could see bruises on his arms. The room was cold from an open window, which had probably been opened hours ago, and yesterday's drinks had not been cleared away--there was no sign that anything had been provided for breakfast. 
Closing the window, Alice noticed a spike of "worry" in her emotion module, directed toward Charlie, whom she knew had a troubled relationship with Oliver, his main caregiver.
The other day, Charlie had acted disruptively due to his late-stage dementia, to which Oliver had expressed frustration and threatened punishment; with his history of crime, substance abuse, unemployment, and mental health problems, this might not be merely an idle threat.
But, there might be some explanation that Alice didn't know about, and she didn't have permission to begin with to contact authorities, since a false report could have highly negative consequences. 
Sending a digital message would also probably not be wise, since the matter was urgent, and Oliver and the rest of the group had access to her logs. 
When she headed over to the reception, there was Oliver talking to Bob.
Alice wanted to let Bob know as soon as possible without alerting Oliver, so she surreptitiously waved to Bob behind Oliver's back to get his attention and flashed a message on her display that she would like to ask him to discreetly check in on Charlie as soon as possible. Bob nodded imperceptibly, and Alice went back to cleaning. With Bob's help, Alice was sure that Charlie would be okay.} 

\begin{figure}[!h]
\centerline{\includegraphics[width=\textwidth]{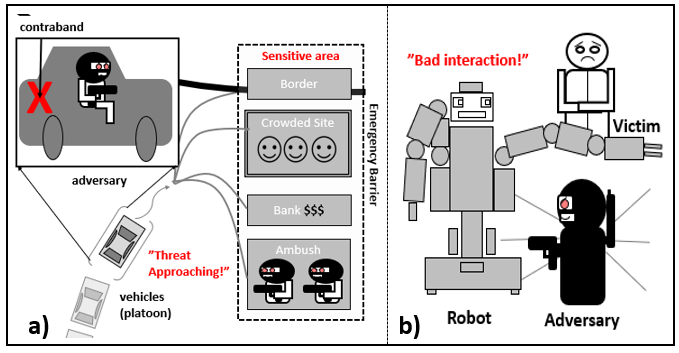}}
\caption{Scenarios: (a) AV. A possible threat is approaching a sensitive area, (b) SAR. Someone might require protection from an undesirable interaction}
\label{fig:scenarios}
\end{figure}

The scenarios suggested that RS might be useful when two conditions hold:
\begin{itemize}
\item {\bf Possible High Danger}. There is a high probability of danger, but the robot is not completely sure about the threat, or has not been given the right to assess such a threat. (The consequences of making a mistake could be extremely harmful and we might not wish to place such power in the hands of a fallible robot.) Thus, the robot requires another opinion, possibly through escalation to a human-in-the-loop.
In particular, this could occur when there is a possibility of an accident or crime.
Traffic accidents are globally the leading killer of people aged 5-29 years, with millions killed and injured annually\footnote{https://www.who.int/publications/i/item/9789241565684}, and crimes are estimated to cost trillions each year~\cite{delisi2016measuring}.
\item {\bf Conventional Communication is Undesirable}. An adversary might detect an unconcealed message, leading to potential reprisals. (People are accustomed to using technologies that could be used against them, from cameras on computers or smart phones that could be used for monitoring, to vehicles that could have a deadly accident. Therefore, an adversary might not want to risk attention by deactivating a robot, or might think it is still useful for their purposes, but if the robot is seen to interfere, the antagonist could try to shut it off, destroy it or abandon it, modify it to either not send messages or send false messages, or find information on the intended recipients; this might make it impossible to help the victim, or punish the victim more, scaling up the problem.) Or, the intended recipient might not have a computer capable of receiving messages, or there might be a lack of connectivity preventing a robot from communicating a problem--with or without covert properties (e.g., because an adversary jammed the wireless channel). 
\end{itemize}
(When these conditions do not hold, a different approach could be used. For example, a robot could directly call for help without steganography if it has witnessed a life-threatening situation and the threat is perfectly clear, like if shots have been fired--or if the robot has a strong belief that the adversary could not detect a call for help.)
A detailed comparison of scenarios for AVs and SARs, including settings, informative cues, and potential robot actions, is presented in Table~\ref{tabDetails}, and the gist is visualized in Fig.~\ref{fig:scenarios}.

\begin{table*}
\caption{Some fundamental concerns for RS with AVs and SARs.}
\label{tabDetails}  
\begin{tabularx}{\linewidth}{ | >{\hsize=0.1\hsize}X | >{\hsize=0.35\hsize}X | >{\hsize=0.55\hsize}X | } \hline
 & 
AVs & 
SARs  \\  \hline
Scenarios & 
Four scenarios for the AV related to potential reckless driving (hit-and-run, drunk driving), trafficking (drugs, humans or other contraband), robbery (at a bank, store, or carjacking), and violent crime (homicide or abduction).  & 
Four scenarios for the SAR related to the potential abduction, abuse, or homicide of vulnerable populations such as elderly, children, persons with special needs (e.g., dementia, autism, blindness, motor impairment,  depression), homeless, spouses, and member of some targeted group (e.g., whistleblowers, freedom fighters, persecuted minorities, or prisoners of war).
\\  \hline
Summary & 
An adversary (lone individual, small group, or representative of an oppressive nation) exhibits malevolent cues during transit to a sensitive area such as a border, bank, military site, or crowded or dangerous location.
The main case involved an adversary travelling inside the AV, but could also include a remotely-controlled AV adversary in a platoon or witnessing an external adversary.

&  
The scenarios mostly involved an adversary that seeks private, sole access to the victim (e.g., via removal to a second location) to avoid abuse being witnessed.
The intention could include domestic violence, kidnapping, bullying, assault, robbery, threat, murder, microaggressions/retribution, battery, rape, or other disturbances--within a setting such as a care center, school, family home, bar/night club, or some other transitional or secluded space such as a street or park. 
The robot could just happen to be in the area or be accompanying a person; such a robot could also be useful for healthcare, daily tasks like cleaning, home assistance, security, or delivery.
\\  \hline
\end{tabularx} 
\end{table*}

\begin{table*}
\caption{Some fundamental concerns for RS with AVs and SARs.}
\label{tabDetails2}  
\begin{tabularx}{\linewidth}{ | >{\hsize=0.1\hsize}X | >{\hsize=0.35\hsize}X | >{\hsize=0.55\hsize}X | } \hline
Cues & 
Cues could include in manual driving mode speeding, weaving, tailgating, and failing to yield or signal, and more generally, hiding packages; unhealthy behavior (medicine non-adherence with depression or sleep deprivation); and being armed and masked without occasion. Problematic driving could be detected by classifying data from surveillance cameras; health problems from collation with data from electronic pill dispensors and smart homes; and threatening gear from cameras inside an AV.

& 
Cues can include: 

(1) Sudden, unexplained, negative or odd changes in a potential victim's state or behavior (possibly due to being drugged, impaired, or intimidated). This can include physical injuries such as bruises, emotion displays of pain, fear, or anger, or payments and subservience to others. As well, this could include worsened environment conditions (e.g., cold and lack of services). To detect such cues, anomaly or change point detection, as well as various health and affective monitoring approaches, could be applied. 

(2) High risk factors such as violent and unjustified behavior or emotional displays from a potential adversary, especially if there is a large perceived force imbalance  (e.g., if a large, armed combatants from a notoriously dangerous group burst in, making violent and unjustified demands), possibly in conjunction with a history of fighting, threats, crime, substance abuse, unemployment, high stress, and mental health problems. Violent behavior could be detected via cameras, microphones, and touch sensors on a SAR or in the environment. If permitted, and cross-application interoperability is enabled, historic data could be accessed from police or medical records.  
  \\  \hline
Actions & 
Warnings could be sent to protective infrastructure (e.g. anti-tire spikes; V2I), border or bank security (V2H), or nearby AVs or platoon members (V2V). &  
Here, the robot’s goal could be to avoid harm to anyone, by discreetly contacting people who might be willing and able to help without the adversary knowing, while preventing the victim from being taken away by possibly lying about the victim’s whereabouts, stalling, delaying, and evading.
Warnings could be sent to family such as parents, security at a bar, teachers, or care center staff.\\  \hline
\end{tabularx} 
\end{table*}

(Along the way, ideas were also considered for two other kinds of robot: A smart home could warn of potential abductions of people held against their will, domestic violence, making/selling/using drugs, or prostitution; also, a wearable suit could warn of a concealed weapon or drugs or some potential crime committed to or by its wearer, such as assault or theft. 
These ideas seemed to be comprised already in the scenarios for AVs and SARs and therefore are not considered in the remainder of the paper.)

\begin{figure}[!t]
\centerline{\includegraphics[width=\textwidth]{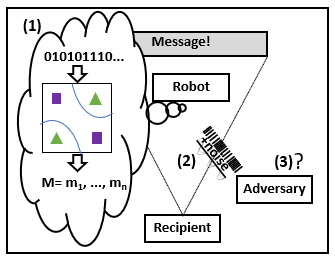}}
\caption{Unique challenges of robot steganography (RS): (1) message generation, (2) indirection, (3) perspective}
\label{fig:unique}
\end{figure}

Additionally, the scenarios also suggested some "unique" aspects to RS that differ from traditional steganography, as shown in Fig.~\ref{fig:unique}:
\begin{itemize}
\item {\bf Generation}.
Instead of humans coming up with messages to send over computers, an AV must itself generate a message from sensed information.
\item {\bf Indirection}.
In HS, files are not passed directly from sender to recipient, introducing risks of noise and lower data transmission rates.
\item {\bf Perspective}.
In HS, unlike e.g. video motion vector steganography, a robot could control its motion to generate an \emph{anisotropic} message visible only by an intended viewer at some specified angle and distance; audio reception could also be controlled via "sound from ultrasound"~\cite{pompei2002sound} or high frequency to send location- or age-specific sounds.
\end{itemize}

\subsection{Carriers}
\label{section:carriers}

As the basic idea of CS is already described in the Section~\ref{section:related-work}, this section focuses
on the new concept of HS and using "physical" carriers:
Visually, locomotion--e.g., variance over time in position and orientation (path, or trajectory), velocity, or acceleration--could be used to hide messages detectable via communicated GPS, videos, or odometry.
Other signals could include lights, and opening or closing of windows and convertible tops.
Aurally, speakers that generate engine noise\footnote{https://www.core77.com/posts/79755/Cars-are-Now-So-Well-Built-Manufacturers-Pipe-In-Fake-Engine-Sounds-Listen-Here}, or even music players or a horn could be used.
More complex approaches could be multimodal, using a platoon, swarm of drones, or even the environment, like birds flying plus an AV’s motion, or use rare modalities like heat; delays, ordering, modality selection, and amplitudes could also be used.

\begin{figure}[!t]
\centerline{\includegraphics[width=\textwidth]{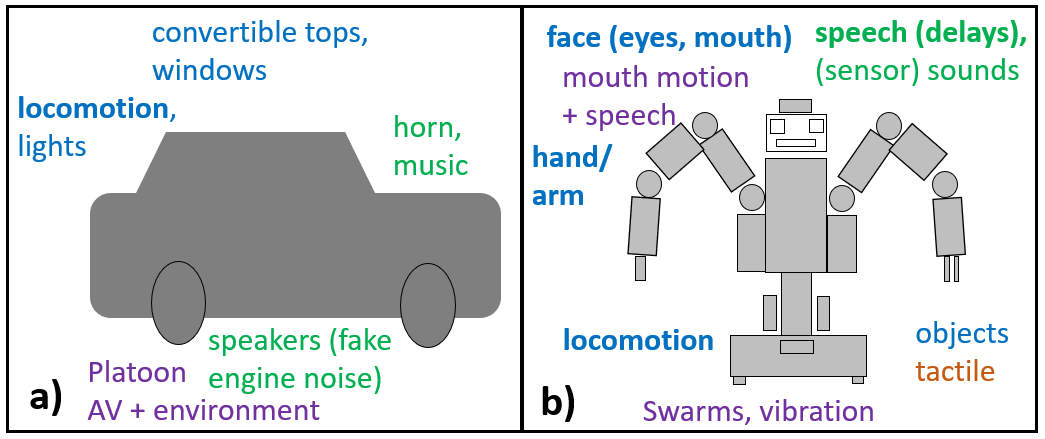}} 
\caption{Carriers: (blue) visual, (green) audio, (purple) multimodal, (brown) other}
\label{fig:carriers}
\end{figure}

\subsection{Signal Generation}
\label{section:signal-generation}

In addition to carriers, how to generate message-bearing signals required consideration.
Various work has looked at how robots can perceive signals and organize knowledge (e.g., based on Semantic Web languages like W3C Web Ontology Language (OWL)~\cite{bruno2019knowledge}), and it is known that information can be encoded in a message as bits or pulses using a code such as ASCII, Morse, or Polybius squares.
What was unclear was how messages can be (1) structured and (2) incorporated in HS (e.g., for locomotion), also from (3) a practical perspective, as considered below:

(1) {\bf Message structuring}.
Assuming that a message consists of 1 to n short propositions $m_i$, of varying time-critical importance $v_i$ (such as the nature of the emergency, location, names, etc.), we propose that message generation can be formulated as a greedily-solvable unbounded Knapsack problem
\begin{equation}
\begin{aligned}
\max \sum v_if(x_i) \\
\mbox{s. t. }T^* = \sum t_ix_i \le T,  \, x_i \ge 0, \, \textrm{and} \\ 
f(2) < \mbox{\textcolor{black}{$2f(1)$}}
\end{aligned}
\label{knapsack_eq}
\end{equation}
where $x_i$ indicates if a proposition $i$ will be included, $t_i$ is the time required to send (not necessarily simply related to message length), $T$ is maximum time available for transmission, and $f$ is an exponential or step function that increases rapidly at first and quickly levels off (e.g.\
\textcolor{black}{
$f\sim e^-{(1/x)}$} for $x\neq0$, $f(0)=0$).

Then, the probability that a proposition is successfully communicated could be calculated as $1-p_i^x$, which is indirectly related to the number of times $x$ that a proposition is sent, where $p_i$ is the probability a given proposition will be "lost" in noisy conditions.

(2) {\bf HS (Locomotion)}.
Assuming a simplified scenario with sideways drifting to encode messages with Morse code, $T$, the time available for transmitting messages can be calculated as $d$, the distance from the front of the AV to its next interruption (e.g., the start of an intersection) divided by the velocity $v_0$ and multiplied by $\alpha$, the desired rate of message to non-message constituents of the carrier signal, related to encoding density (e.g.$1:10$)
\begin{equation}
T \;=\;\mbox{\textcolor{black}{$\displaystyle\alpha\cdot \frac{d}{v_0}$}}
\end{equation}

Given knowledge of what to send and when, we next calculate how a robot's motions could be controlled.
First, a motion generation model is required.
Various such models exist; here we explore a simplified extension of the circular specification of the social force model, that can be trivially extended to provide a simple model for a robot's motion~\cite{helbing2013pedestrian, cooney2012designing}.
Nearby lanes/barriers, passing vehicles, and cordoned off road sections exert mainly a lateral force on the vehicle, whose forward speed is maintained within the speed limit; whereas close red light intersections and stopped cars in front exert a frontal force\textcolor{black}{, as expressed by}
\begin{eqnarray}
\textcolor{black}{\vec{f_{\rm net}}} &=& 
      \textcolor{black}{\vec{f_{\rm goal}}} + \textcolor{black}{\vec{f_{\rm env}}} + \textcolor{black}{\vec{f_{\rm com}}}\\[3mm] 
\textcolor{black}{\vec{f_{\rm sf}}} &=& \sum A_ie^{-\textcolor{black}{d_{\rm ro}}/B_i}\textcolor{black}{ \frac{\vec{d_{\rm ro}} }{d_{\rm ro}}    } 
\end{eqnarray}
where $\vec{f_{\rm goal}}$ is a force acting on the robot to ensure it moves toward its goal in an appropriate way, maintaining a steady forward speed that obeys the speed limit,
$\vec{f_{\rm env}}$ represents forces exerted by the nearest environment (lanes, obstacles, robots, traffic lights, and other infrastructure), and 
$\vec{f_{\rm com}}$ is a force working on the robot to communicate some hidden message.
(Other terms could also be used for more complex modelling, such as a social filter to make the robot's intentions clearer to human drivers, or some random noise when not transmitting meaningful messages intended to make it harder for adversaries to detect messaging (e.g., "salt": a sequence of random meaningless bits to be concatenated with the information bearing sequence).
As well, PID controllers can be used for all force terms, to ensure that the robot moves correctly.)

\textcolor{black}{In (4), the variable} $\vec{d_{\rm ro}}$ is the 3D displacement vector from the robot’s
center of mass to an obstacle’s center of mass ($d_{\rm ro}$ is the
magnitude of such a vector, or relative distance), $A$ is the
“interaction force”, and $B$ is the “interaction length” a robot 
seeks to establish between itself and obstacles.
At stops, the force could also be calculated so that the robot's speed becomes zero rather than moving backwards, if $\textcolor{black}{\sin}\, \theta < 0$, where $\theta$ is the angle between the lateral \textcolor{black}{$x$} axis and the robot's velocity vector.

Also, to cause a robot to drift slightly sideways, a similar equation could be used, by "hallucinating" the presence of a "virtual obstacle" in front and to the side of the robot.
Thus, such a model, in conjunction with standard (Ackermann) kinematics and dynamics, could be used to calculate the parameters required for a robot to combine its required motion and steganographic motion.

(3) {\bf Practical concerns}.
Reality can also raise its "ugly head" when designers seek to move from theory to practical applications.
Table \ref{tabPractical} summarizes practical constraints related to frequency, accuracy, and potential challenges.\footnote{1cm is our estimate based on a typical dashcam and 10m distance, and 1mm is reported in: https://www.manufacturingtomorrow.com/article/2018/01/industrial-robots-encoders-for-tool-center-point-accuracy/10867/}

\begin{table}
\caption{Some practical considerations for potentially useful carriers.}
\label{tabPractical}  
\begin{tabularx}{\linewidth}{ | >{\hsize=0.2\hsize}X | >{\hsize=0.1\hsize}X | >{\hsize=0.4\hsize}X | >{\hsize=0.3\hsize}X | } \hline
1 GPS           & 5-10Hz   & 5m (direct) 30-50 cm (DGPS), 2cm (RTK)~\cite{perez2012tractor} & How to get data (CAM)?     \\ \hline
2 Video         & 30 fps &  $>$1cm (relative), 1mm (indoor, markers) & Weather \\ \hline
3 Audio onset & 40Hz  & $F_1$= 0.817  & Noise  \\ \hline
\end{tabularx} 
\end{table}

The details for our estimate regarding relative positioning via camera are as follows.
If it is imagined that one robot is watching another robot in front, how much would the front robot have to move laterally for the motion to be visible by the back robot?
For example, if we assume a standard $1920\textcolor{black}{\times}1080$ dashcam (the total resolution horizontally $\mbox{res}_{\rm tw}$ is 1920), with a \textcolor{black}{viewing angle}, $\phi$, of 130 degrees, 10 m distance between robots $d_{ij}$, the smallest resolution of motion that can be detected for one pixel to change $d_{\rm pix}$ is approximately 1 cm of relative movement, as expressed by
\begin{equation}
    \textcolor{black}{d_{\rm pix}} = \frac{2\tan^{-1} (\frac{\phi}{2})}  {d_{ij} \mbox{res}_{\rm tw}} 
\end{equation}

Thus, as noted, various challenges exist--from accurate perception and control, to the numerous other channels that could be utilized for HS, whose exploration lies outside of the scope of this speculative paper (it is not the goal of this paper to fully solve this problem, but rather to explore it and stimulate discussion). 
Nonetheless, the rich AI and robotics literature appears to contain approaches that could be adapted to tackle robot signal generation, also for HS.

\section{User Study}
\label{section:user-study}

Various proposals were made, but can messages truly be hidden in a robot's motions and sounds in a way that is not easy to detect by potential adversaries?
To check the basic feasibility of this idea, a small user study was conducted based on implementing some simplified algorithms.
We wished to develop an actual robot prototype rather than a simulation where real-world problems might not emerge; since an error with an AV could be lethal, we selected a SAR to use for our first exploration.

\subsection{Participants}

20 faculty members and students at our university's School of Information Technology participated in an online survey (40\% were female, 50\% male, and 10\% preferred not to say; average age was 41.8 years with SD= 10.3; and six nationalities were represented, with Swedish by far the most common (60\%)). Participants received no compensation.
Ethical approval was not required for this study in accordance with the Swedish ethics review act of 2003 (SFS no 2003:460), but principles in the Declaration of Helsinki and General Data Protection Regulation (GDPR) were followed: e.g., in regard to written informed consent.

\subsection{Procedure}

Similar to our previous study on audio steganography~\cite{jarpe2021velody}, participants were sent links to a Google Forms survey, which took approximately three minutes to complete.
Participants watched two pairs of videos of a robot moving and speaking.
In one video of each pair, messages were hidden in the robot’s behaviors. 
After watching each pair of videos, participants noted which one they thought contained a hidden message.
For safety and simplicity, the Baxter robot was used, which is a SAR that is easily programmed to communicate audiovisually via movements, a face display, and speech utterances.

\subsection{Conditions}

\begin{figure}[!t]
\centerline{\includegraphics[width=\textwidth]{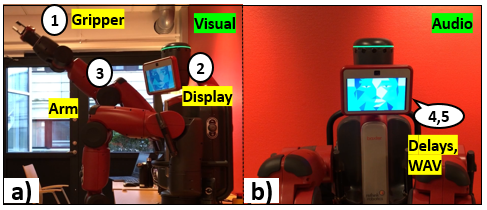}}
\caption{Audiovisual steganography was applied to (a) visual signals (gripper motions, facial display, arm motions), as well as (b) sounds (speech delays and WAV files).  
For a better understanding of the signals, we refer the reader to the accompanying video for this paper (youtu.be/vr3zlva6cCU).}
\label{fig:exp}
\end{figure}

There were two conditions, steganography (present or absent) and modality (visual or audio).
Thus, four videos (two pairs) were prepared, as depicted in Fig.\ref{fig:exp}.
Both of the steganography videos featured both examples of CS and HS.

Visual steganography: The first two videos show the robot greeting an adversary who has suddenly appeared. (The videos are recorded from the adversary's perspective.)
The steganography video uses three visual carriers to send hidden messages to an off-camera observer to the left or front of the robot, intended to be difficult to perceive by an adversary to the robot's right.
The robot initially appears as if it is vacuum cleaning, while opening and closing a gripper to express "SOS" through Morse code; the gripper's motion and clicking sound are masked via its other arm--demonstrating the perspective property--and the vacuuming sound.
Next, the robot raises its arm in greeting, during which arm angles specified as floats are recorded in a motion file and constantly published on Robot Operating System (ROS) channels; there, an ASCII message "SOS" was embedded in the LSB. 
Also, the message "SOS! Stacey needs your help!" was hidden in the image data used for the robot's face via LSB/ASCII.
Thus, the messages are conveyed via "pure" HS (not requiring a computer to detect), HS closer to CS (a human-like motion, but requiring a computer), and CS.
It wasn't clear if participants might notice strangeness in the clicking of the gripper, vibrations due the concealed motion, or from roughness in the robot's arm motion due to the noise added.

Audio steganography: 
The second pair of videos shows a robot speaking to an adversary.
The steganography video uses two audio carriers to hide messages.
Speech utterance delays were used to hide the message "SOS" in simplified Morse code (100ms dot, 200ms dash).
The message is read by using the sounddevice library to detect volume, and a model of normal delays--exemplifying the indirection property. Some simple heuristics were used including hysteresis in volume for speech start and end, and a model for minimal length of a speech sound.
Furthermore, the message "SOS! Stacey needs your help!" was again hidden, in a WAV file the robot uses to say "hello", via LSB/ASCII.
Thus, the signals are predominantly HS and CS, respectively. 
It wasn't clear if participants might notice the delays, or roughness in the sounds.

Note: We claimed at the start of this paper that encryption alone cannot hide that some transmission is taking place, but why then can we say that transmission is hidden in a case like this? The reason is that an encrypted message is clearly a message that the sender does not want read; whereas, we expect a SAR to use its display, motions, and sounds to interact--normally there is no suspicion that such behaviors have any other hidden communicative purpose.

\subsection{Results}

If the robot was successful at hiding messages, the rate at which participants could identify the videos with hidden messages should be around random chance (50\%).
If the robot failed to hide its messages, the rate should be higher (e.g., closer to 100\%).

In this study, participants guessed the file with the hidden messages 55\% and 50\% of the time (11 and 10 participants respectively; thus, 21 out of 40 times, or 52.5\%).
A binomial test confirmed that detection was not more common than can be expected by random chance (50\%), for the results of the first pair of videos ($p\!=\!.8$), the second pair ($p\!>\!.9$), and all of the data together ($p\!=\!.9$). 
In other words, the robot had performed as we had expected.
Also, of 13 participants who left comments, ten stated that they could not see obvious differences and guessed; three indicated cues such as nodding, color changes, pauses and blinking that could not be clearly related to changes expected due to embedding messages, possibly a result of \emph{pareidolia} (a human tendency to see patterns even where none exist).

Thus, our hypothesis was supported, and this simplified check confirmed that messages can be successfully concealed in some common robot behaviors without humans suspecting. (While some messages require the recipient to have a computer, others, like the robot's gripper communicating via Morse code, can be seen by the naked eye.)

\section{Discussion}
\label{section:discussion}

In summary, the contribution of the current work is proposing some theoretical and practical considerations for a robot to convey hidden messages to help people, which we have dubbed \emph{robot steganography} (RS):

\begin{itemize}
\item{A speculative approach revealed}
\begin{itemize}
\item{applications to traffic safety and crime prevention}
\item{three unique qualities of RS relating to message generation, indirection, and perspective}
\item{potential carriers, as well as initial ideas for signal generation, comprising message structuring, motion generation, and some practical constraints.}
\end{itemize}
\item{A simplified user study confirmed that messages can be hidden in various robot behaviors, also demonstrating a first example of robot steganography.}
\item{Additionally, a video and code have been made freely available to help guide others who might be interested in this topic.\footnote{youtu.be/vr3zlva6cCU}\footnote{github.com/martincooney/robot-steganography}.}
\end{itemize}

The immediate implication is that robot designers might wish to be aware of the idea of RS, which could save people's lives and could be easily implemented in some contexts:
Robots can already use established approaches for computer-based steganography (CS) to communicate with computers, robots, or humans equipped with computers.
Moreover, human-like RS (HS) can also be used to communicate even with humans who do not have access to a computer (e.g., this can be as simple as merely gesturing or displaying a message behind someone's back), when technically capable adversaries might be aware of more common steganography approaches, or when conventional CS channels are obstructed.
Furthermore, as noted, RS can complement other techniques such as encryption or lying. For example, while sending messages, a robot could temporarily pretend to slow down or turn, to fool an adversary who wants to stop or guide it toward a threat.

\subsection{Limitations and Future Work}

This work represents only an initial investigation into a few topics related to RS and is limited by the speculative prototyping approach followed. 
It should be stressed that it is neither possible, nor the goal of the current paper, to fully describe any and all kinds of RS methods that could exist. 
Other important scenarios, carriers, and motion generation approaches might exist for other kinds of robots, and the simplified user study only checked the feasibility of RS in a prototype SAR, since a mistake in an AV could be lethal.
As well, the usefulness of RS is not completely clear, as CS can result in lower transmission rates than merely encrypting: for HS, there can be further restrictions on speed and robustness, which might not be justified if intended recipients are expected to always be monitoring a computer or adversaries are expected to never be technically capable.

Future work will explore steganalysis (detection of hidden messages), ethics, and perception of dangerous scenarios:
\begin{itemize}
\item{ {\bf Robot Steganalysis} (RS2).
How can threats be detected if recipients must look at massive amounts of data?
Discrepancies from a "normal" model might be undesirable, since they could be detected also by adversaries, but mimic functions and data whitening could complicate detection.
One basic steganalysis approach involves comparing signals known to be good with suspect signals, assuming adversaries do not have such data; and, decentralized systems and federated detection could be useful.

Another related concern is if adversaries can neutralize messaging (\emph{content threat removal}) via "jamming"?--For example, by shining light, occluding, or playing loud sounds.
A countermeasure could be to use robust techniques (e.g., not LSB), large signals (e.g., large motions like turning could be easier to hide messages), and multimodal redundancy (if sound is jammed, maybe a visual message might still pass) and repetition of messages.
Conversely, if HS is too easily detected, a robot could spoof being piloted by a human; e.g., it might be strange if an AV weaves around the road, but could be overlooked if it is believed the robot is being piloted by a tired, unskilled, or playful human driver.

Steganalysis could also be used to detect infiltrated "Byzantine" systems. For example, a robot in an insecure platoon could use a "canary trap", sending some watermarked messages to seek to detect an adversary: thus, watermarking steganography to improve authenticity.
Conversely, where anonymity is a concern, a robot could use a Man in the Middle (MITM) approach to capture an adversary’s or other communications (like CAM), embedding a hidden message and propagating it, to make it harder for the antagonist to identify the robot as the sender of a hidden message.}
\item{ {\bf Ethics}.
Future work must also examine the ethical challenges in RS and propose solutions. 
New technologies are increasingly becoming misused by nefarious agents such as terrorists and hackers--it is the age of Deepfakes and alternative truth in media, of trolls and botnets, of the Dark Web and malicious Stegware code. 
Here, adversaries could reprogram robots to send secret messages about potential targets. Can we trust a system that could betray us, or make mistakes?--We believe the answer could be yes, because we trust humans who wield similar power over us.
Thus, also given the typical "arm's race" that takes place between defenders and adversaries, our aim is that such knowledge should not allowed to be solely in the possession of nefarious agents, to prevent “Black Mirror”-like future scenarios; however, it is clear that such potential problems should be carefully examined.}
\item{ {\bf Perception}.
Perception of potentially dangerous cues, such as injuries or weapons, is non-trivial, especially if people's lives and reputations might be at stake. This also relates to localization and a robot's theory of mind, including inference of what an adversary or intended recipient can see or hear. Although extremely challenging, excellent inferential abilities from subtle cues that in humans are typically attributed to detectives or spies could also be useful.}
\end{itemize}

By shining light on such questions that seems to have not yet received much attention, we aim to bring a fresh perspective to possibilities for robots to create a better, safer society, which could also facilitate acceptance and trust in the use of robots in our everyday lives.

\section*{Acknowledgment}

We acknowledge the kind help of our colleagues who participated in scenario building or volunteered as participants to help check our prototype!

\bibliographystyle{IEEEtranS}
\bibliography{cooney_robot_steganography_arxiv}

\begin{thebibliography}{10}
\providecommand{\url}[1]{#1}
\csname url@samestyle\endcsname
\providecommand{\newblock}{\relax}
\providecommand{\bibinfo}[2]{#2}
\providecommand{\BIBentrySTDinterwordspacing}{\spaceskip=0pt\relax}
\providecommand{\BIBentryALTinterwordstretchfactor}{4}
\providecommand{\BIBentryALTinterwordspacing}{\spaceskip=\fontdimen2\font plus
\BIBentryALTinterwordstretchfactor\fontdimen3\font minus
  \fontdimen4\font\relax}
\providecommand{\BIBforeignlanguage}[2]{{%
\expandafter\ifx\csname l@#1\endcsname\relax
\typeout{** WARNING: IEEEtranS.bst: No hyphenation pattern has been}%
\typeout{** loaded for the language `#1'. Using the pattern for}%
\typeout{** the default language instead.}%
\else
\language=\csname l@#1\endcsname
\fi
#2}}
\providecommand{\BIBdecl}{\relax}
\BIBdecl

\bibitem{bruno2019knowledge}
B.~Bruno, C.~T. Recchiuto, I.~Papadopoulos, A.~Saffiotti, C.~Koulouglioti,
  R.~Menicatti, F.~Mastrogiovanni, R.~Zaccaria, and A.~Sgorbissa, ``Knowledge
  representation for culturally competent personal robots: requirements, design
  principles, implementation, and assessment,'' \emph{International Journal of
  Social Robotics}, vol.~11, no.~3, pp. 515--538, 2019.

\bibitem{cooney2021}
M.~Cooney, E.~Järpe, and A.~Vinel, ````vehicular steganography''?:
  Opportunities and challenges,'' in \emph{Short extended abstract submitted to
  the Nets4Cars workshop}, 2021.

\bibitem{cooney2012designing}
M.~Cooney, F.~Zanlungo, S.~Nishio, and H.~Ishiguro, ``Designing a flying
  humanoid robot (fhr): effects of flight on interactive communication,'' in
  \emph{2012 IEEE RO-MAN: The 21st IEEE International Symposium on Robot and
  Human Interactive Communication}.\hskip 1em plus 0.5em minus 0.4em\relax
  IEEE, 2012, pp. 364--371.

\bibitem{de2014applying}
J.~M. de~Fuentes, J.~Blasco, A.~I. Gonz{\'a}lez-Tablas, and
  L.~Gonz{\'a}lez-Manzano, ``Applying information hiding in vanets to covertly
  report misbehaving vehicles,'' \emph{International Journal of Distributed
  Sensor Networks}, vol.~10, no.~2, p. 120626, 2014.

\bibitem{delisi2016measuring}
M.~DeLisi, ``Measuring the cost of crime,'' \emph{The handbook of measurement
  issues in criminology and criminal justice}, pp. 416--33, 2016.

\bibitem{dunne2013speculative}
A.~Dunne and F.~Raby, \emph{Speculative everything: design, fiction, and social
  dreaming}.\hskip 1em plus 0.5em minus 0.4em\relax MIT press, 2013.

\bibitem{engelberg2002framework}
D.~Engelberg and A.~Seffah, ``A framework for rapid mid-fidelity prototyping of
  web sites,'' in \emph{IFIP World Computer Congress, TC 13}.\hskip 1em plus
  0.5em minus 0.4em\relax Springer, 2002, pp. 203--215.

\bibitem{10.3389/frobt.2019.00064}
\BIBentryALTinterwordspacing
J.~Gonzalez-Billandon, A.~M. Aroyo, A.~Tonelli, D.~Pasquali, A.~Sciutti,
  M.~Gori, G.~Sandini, and F.~Rea, ``Can a robot catch you lying? a machine
  learning system to detect lies during interactions,'' \emph{Frontiers in
  Robotics and AI}, vol.~6, p.~64, 2019. [Online]. Available:
  \url{https://www.frontiersin.org/article/10.3389/frobt.2019.00064}
\BIBentrySTDinterwordspacing

\bibitem{helbing2013pedestrian}
D.~Helbing and A.~Johansson, ``Pedestrian, crowd, and evacuation dynamics,''
  \emph{arXiv preprint arXiv:1309.1609}, 2013.

\bibitem{isaac2017robots}
A.~M. Isaac and W.~Bridewell, ``Why robots need to deceive (and how),''
  \emph{Robot ethics}, vol.~2, pp. 157--172, 2017.

\bibitem{jarpe2021velody}
E.~J{\"a}rpe and M.~Weckst{\'e}n, ``Velody 2—resilient high-capacity midi
  steganography for organ and harpsichord music,'' \emph{Applied Sciences},
  vol.~11, no.~1, p.~39, 2021.

\bibitem{jia2018bio}
J.~Jia-jia, W.~Xian-quan, D.~Fa-jie, F.~Xiao, Y.~Han, and H.~Bo, ``Bio-inspired
  steganography for secure underwater acoustic communications,'' \emph{IEEE
  Communications Magazine}, vol.~56, no.~10, pp. 156--162, 2018.

\bibitem{ju2015design}
W.~Ju, ``The design of implicit interactions,'' \emph{Synthesis Lectures on
  Human-Centered Informatics}, vol.~8, no.~2, pp. 1--93, 2015.

\bibitem{perez2012tractor}
M.~Perez-Ruiz, D.~C. Slaughter, C.~Gliever, and S.~K. Upadhyaya,
  ``Tractor-based real-time kinematic-global positioning system (rtk-gps)
  guidance system for geospatial mapping of row crop transplant,''
  \emph{Biosystems engineering}, vol. 111, no.~1, pp. 64--71, 2012.

\bibitem{petitcolas1999information}
F.~A. Petitcolas, R.~J. Anderson, and M.~G. Kuhn, ``Information hiding-a
  survey,'' \emph{Proceedings of the IEEE}, vol.~87, no.~7, pp. 1062--1078,
  1999.

\bibitem{pompei2002sound}
F.~J. Pompei, ``Sound from ultrasound: The parametric array as an audible sound
  source,'' Ph.D. dissertation, Massachusetts Institute of Technology, 2002.

\bibitem{rasmussen2005narrative}
L.~B. Rasmussen, ``The narrative aspect of scenario building-how story telling
  may give people a memory of the future,'' \emph{AI \& society}, vol.~19,
  no.~3, pp. 229--249, 2005.

\bibitem{schmidt2000implicit}
A.~Schmidt, ``Implicit human computer interaction through context,''
  \emph{Personal technologies}, vol.~4, no.~2, pp. 191--199, 2000.

\bibitem{wagner2016lies}
A.~R. Wagner, ``Lies and deception: Robots that use falsehood as a social
  strategy,'' \emph{Robots that talk and listen: Technology and social impact.
  De Grutyer https://doi. org/10.1515/9781614514404}, 2016.

\bibitem{wang2020intelligent}
R.~Wang, R.~Sell, A.~Rassolkin, T.~Otto, and E.~Malayjerdi, ``Intelligent
  functions development on autonomous electric vehicle platform,''
  \emph{Journal of Machine Engineering}, vol.~20, 2020.

\bibitem{zielinska2014trends}
E.~Zieli{\'n}ska, W.~Mazurczyk, and K.~Szczypiorski, ``Trends in
  steganography,'' \emph{Communications of the ACM}, vol.~57, no.~3, pp.
  86--95, 2014.

\end{thebibliography}

\vspace{12pt}

\end{document}